\documentclass[10pt, letterpaper]{article}   	
\usepackage{opex3}
\usepackage{color}
\usepackage{float} 					
\usepackage{amssymb}
\usepackage{amsmath}
\usepackage{textcomp}
\usepackage{xspace}
\usepackage{multirow}
\usepackage{cite}
\newcommand*{\eg}{e.g.\@\xspace}
\newcommand*{\ie}{i.e.\@\xspace}
\DeclareMathOperator*{\argmin}{\arg\!\min}

\begin{document}

\title{Compressive Holographic Video}

\author{Zihao Wang$^1$$^*$, Leonidas Spinoulas$^1$, Kuan He$^1$, Huaijin Chen$^2$,\\ Lei Tian$^3$, Aggelos K. Katsaggelos$^1$ and Oliver Cossairt$^1$}

\address{$^1$Department of Electrical Engineering and Computer Science, Northwestern University \\ Evanston, IL 60208, USA\\$^2$Department of Electrical and Computer Engineering, Rice University \\ Houston, TX 77005,USA\\$^3$Department of Electrical and Computer Engineering, Boston University \\Boston, MA 02215, USA}

\email{$^*$zwinswang@gmail.com}


\begin{abstract}
Compressed sensing has been discussed separately in spatial and temporal domains. Compressive holography has been introduced as a method that allows 3D tomographic reconstruction at different depths from a single 2D image. Coded exposure is a temporal compressed sensing method for high speed video acquisition. In this work, we combine compressive holography and coded exposure techniques and extend the discussion to 4D reconstruction in space and time from one coded captured image. In our prototype, digital in-line holography was used for imaging macroscopic, fast moving objects. The pixel-wise temporal modulation was implemented by a digital micromirror device. In this paper we demonstrate $10\times$ temporal super resolution with multiple depths recovery from a single image. Two examples are presented for the purpose of recording subtle vibrations and tracking small particles within 5 ms.
\end{abstract}

\ocis{(090.1995) Digital holography; (110.1758) Computational imaging.}

\section{Introduction}
\label{sec:intro}
In recent years, people have witnessed a great interest in exploiting the redundant nature of signals. The redundancy of acquired signals provides the opportunity to sample data in a compressive approach. Cand{\`e}s et al.~\cite{Candes2006,Candes2006b} and Donoho~\cite{Donoho2006} have discussed the high probability of reconstructing signals with high fidelity from few random measurements, provided that they are sparse or compressible in a transformation basis. Since then, the theory of compressed sensing (CS) has been widely applied to computational imaging. Lustig et al.~\cite{Lustig2008} described the natural fit of CS to magnetic resonance imaging (MRI). Gan~\cite{Gan2007} proposed block compressed sensing method for natural images, which is applicable for low-power, low-resolution imaging devices. Brady et al.~\cite{Brady2009} showed that holography can be viewed as a simple spatial encoder for CS and demonstrated 3D tomography from 2D holographic data.

Gabor's invention of holography in 1948~\cite{Gabor1948} has provided an effective method for recording and reconstructing a 3D light field from a captured 2D hologram. The use of a CCD camera to digitally record holographic interference patterns has made digital holography (DH) an emerging technology with a variety of imaging applications, such as particle imaging, tracking in biomedical microscopy~\cite{Memmolo2015,Xu2001,Su2012, Lu2009,Dixon2011} and physical process profiling and measuring~\cite{Saxton1997, Katz2010, Tian2010, Xu2003, Nilsson2000}.  Digital Gabor/in-line holography (DIH) is a simple, lensless, yet effective setup for capturing holograms. The simplicity of DIH is balanced by the requirement that objects be small enough to avoid occluding the reference beam significantly~\cite{Kim2011}. Extensive discussions and applications of DIH have been focused on microscopic imaging, \ie small and fast-moving objects~\cite{Garcia-Sucerquia2006, Xu2001b, Yu2014b}. The tracking of fast movements usually entails multiple exposures~\cite{Memmolo2015, Su2012, Tian2010, Xu2003, Salah2008, Yu2014}. Temporal resolution is usually limited to the 10-100 millisecond range and little research has been conducted on temporal compression. However, in recent years, CS has proved a useful tool to increase the spatial information encoded in DH~\cite{Brady2009,Lim2011}. Rivenson et al.~\cite{Rivenson2010} discussed the application of CS to digital Fresnel holography. Liu et al.~\cite{Liu2012} and Song et al.~\cite{Song2016} improved subpixel accuracy for object localization and enhanced spatial resolution (super-resolution). Furthermore, CS theory has proven successful for recovering scenes under holographic microscopic tomography~\cite{Hahn2011}, off-axis frequency-shifting holography~\cite{Marim2010} as well as millimeter-wave holography~\cite{Cull2010}. Coded apertures have also been used together with CS to provide robust solutions for snapshot phase retrieval~\cite{Horisaki2014, Egami2016}. In view of recent research in CS and DH, several natural questions arise: Can we extend coded aperture to coded exposure? Can we exploit the unused pixels in exchange for increased temporal resolution? Since holography is naturally suitable for recovering depth information, a further research question is whether 4D space-time information can be extracted from 2D data employing the CS framework.

Similar discussions have been initiated in the incoherent imaging regime. Leveraging multiplexing schemes in the temporal domain, \eg coded exposure, has been demonstrated as an effective hardware strategy for exploiting spatiotemporal trade-offs in modern cameras. High speed sensors usually require high light sensitivity and large bandwidth due to their limited on-board memory. In 2006, Raskar et al.~\cite{Raskar2006} pioneered the concept of coded exposure when he introduced the flutter shutter camera for motion deblurring. The technique requires knowledge of motion magnitude/direction and cannot handle general scenes exhibiting complex motion. Bub et al.~\cite{Bub2010} designed an amphibious (both microscopic and macroscopic) high speed imaging system using a DMD (digital micromirror device) for temporal pixel multiplexing. Gupta et al.~\cite{Gupta2010} showed how per-pixel temporal modulation allows flexible post-capture spatiotemporal resolution trade-off. Reddy et al.~\cite{Reddy2011} used sparse representations (spatial) and brightness constancy (temporal) to preserve spatial resolution while achieving higher temporal resolution. Liu et al.~\cite{Liu2014} used an over-complete dictionary to sparsely represent time-varying scenes. Koller et al.~\cite{Koller2015} discussed several mask patterns and proposed a translational photomask to encode scene movements extending the work of~\cite{Llull2013}. These methods have proved successful for reconstructing fast moving scenes by combining cheap low frame-rate cameras with fast spatio-temporal modulating elements. While all of these techniques enable high speed reconstruction of 2D motion, incorporating holographic capture offers the potential to extend the capabilities to 3D motion. Moreover, in many holography setups, the energy from each scene is distributed across the entire detector so that each pixel contains partial information about the entire scene. This offers the potential for improved performance relative to incoherent architectures.

Our work exploits both spatial and temporal redundancy in natural scenes and generalizes to a 4D (3D positon with time) system model. We show that by combining digital holography and coded exposure techniques using a CS framework, it is feasible to reconstruct a 4D moving scene from a single 2D hologram. We demonstrate $10\times$ temporal super resolution and anticipate optical sectioning for 1 cm. As a test case, we focus on macroscopic scenes exhibiting fast motion of small objects (vibrating bars or small particles, etc.).

\section{Generalized system model}
Digital Gabor holography requires no separation of the reference beam and the object beam. The object is illuminated by a single beam and the portion that is not scattered by the object serves as the reference beam. This concept leads to a simple experimental setup but demands limited object sizes so that the reference beam is not excessively disturbed. In this case, the imaging process is a recording of the diffraction pattern of a 2D aperture.

\subsection{Diffraction theory}
We first model diffraction in a 2D aperture case. According to Fresnel-Kirchoff diffraction formula~\cite{Kim2011}, the field at each observation point $E(x, y; z)$ in a 2D aperture can be written as
\begin{equation} \label{eq:propagation}
\begin{split}
E(x, y; z) & = -\frac{ik}{2\pi}\iint_{\Sigma_0}E_0(x_0, y_0)\frac{\exp(ikr)}{r}\,dx_0\,dy_0,\\
& =-\frac{ik}{2\pi z}\iint_{\Sigma_0}E_0(x_0, y_0)\exp\left\{ik\left[(x-x_0)^2 + (y-y_0)^2 + z^2\right]^{\frac{1}{2}}\right\},
\end{split}
\end{equation}
where $r$ denotes the distance from $(x_0, y_0)$ at the input plane $\Sigma_{0}$,with input field $E_{0}(x_0, y_0)$, to $(x, y)$ at the output plane, \ie $r = \left[(x-x_0)^2 + (y-y_0)^2 + z^2\right]^{\frac{1}{2}}$. In the second line of Eq. \eqref{eq:propagation} we make a further approximation of $r \approx z$ in the denominator, but not in the exponent. The integral then becomes a convolution,
\begin{equation} \label{eq:convolution}
E(x, y; z) = H \ast E_0 ,
\end{equation}
with the kernel
\begin{equation} \label{eq:kernel}
H(x, y; z) = -\frac{ik}{2\pi z}\exp\left[ik\left(x^2 + y^2 + z^2\right)^{\frac{1}{2}}\right].
\end{equation}

The kernel $H$ is also referred to as the point spread function (PSF). Since the propagation is along the $z$-axis, the form of the kernel is determined by the propagation distance $z$.
\begin{figure}[!t]
	\centering\includegraphics[width=4.5in,height=1.98in]{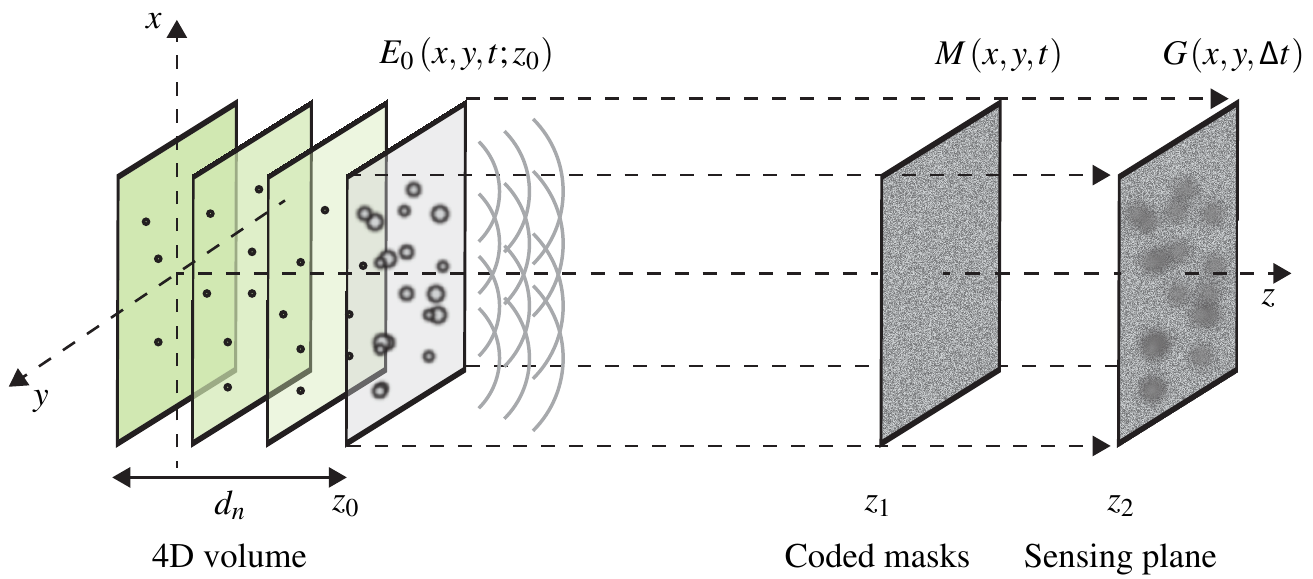}
	\caption{4D holographic model. $E_0(x, y, t; z_0)$: projection of a 4D field at $z_0$, the $n$-th depth plane has a distance of $d_n$ to $z_0$; $M(x, y, t)$: temporal coded mask located at $z_1$; $G(x, y, \Delta t)$: captured image with an integral over $\Delta t$. The sensor is located at $z_2$.}
	\label{fig:4D_model}
\end{figure}
\subsection{4D model}
\label{subsec:4D_model}
We now extend our analysis to a 4-dimensional model. As illustrated in Fig.~\ref{fig:4D_model}, consider a 4D field $V(x, y, z, t)$, which propagates along the positive $z$-direction. Along the propagation path, a high-speed coded mask $M(x, y, t)$ is located at $z_1$. A sensor is placed on the sensing plane $z_2$. In one frame, the sensor captures the intensity of the field during an exposure time of $\Delta t$. The volume can be discretized into $N_d$ planes, with the furthest plane having a distance of $d_n$ with respect to the observation plane at $z_0$. In Gabor holography, the object beam and the reference beam overlap with each other. This requires the objects to be sparse so that the occlusion of the reference beam is negligible. Under this assumption, the field $V$ in reality represents the summation of the object field and the constant reference field. Thus, the field at $z_0$ is 
\begin{equation} 
\label{eq:volume}
E_0(x, y, t; z_0) = \sum_{n=1}^{N_d} H_{d_n} \ast O(x, y, z_0-d_n, t) + R,
\end{equation}
where $H_{d_n}$ denotes the convolutional kernel for distance $d_n$. 

At the sensing plane, during one exposure time $\Delta t$, the sensed image can be expressed as an integral of the intensity $I$ of the field as
\begin{equation} \label{eq:time}
\begin{split}
G(x, y, \Delta t) & = \int_{t=t_0}^{t_0+\Delta t} I(x, y, t; z_2) dt\\
& = \int_{t=t_0}^{t_0+\Delta t} \left\vert H_{z_2-z_1} \ast\Big\{M(x, y, t)\left[H_{z_1-z_0} \ast E_0(x, y, t; z_0)\right]\Big\} \right\vert ^2 dt.
\end{split}
\end{equation}

Equation~\eqref{eq:time} describes the continuous form of the sensing process. However, the mask is operated in a discrete form at high frame rates. Suppose that for each sensor frame the coded mask changes $T$ times at equal intervals of $\tau = \Delta t/T$ during $\Delta t$. Then the discretized form of $G$ is 
\begin{equation} \label{eq:discreteG}
\begin{split}
G(x, y, \Delta t) & = \sum_{i=0}^{T-1} \int_{t=t_i}^{t_{i+1}}\left\vert H_{z_2-z_1} \ast \Big\{M(x, y, t)\left[H_{z_1-z_0} \ast E_0(x, y, t; z_0)\right]\Big\}\right\vert^2dt\\
& = \tau\sum_{i=0}^{T-1} \left\vert H_{z_2-z_1} \ast \Big\{M(x, y, t_i) \left[H_{z_1-z_0} \ast E_0(x, y, t_i; z_0)\right]\Big\}\right\vert^2\\
& = \tau\sum_{i=0}^{T-1} \left\vert H_{z_2-z_1} \ast \left\{M(x, y, t_i) \left[H_{z_1-z_0} \ast \left(\sum_{n=1}^{N_d} H_{d_n}\ast O(x, y, z_0-d_n, t) + R \right)\right]\right\}\right\vert^2\\
& = \tau\sum_{i=0}^{T-1} \left\vert O_{c, i} + R_{c, i} \right\vert^2,
\end{split}
\end{equation}
where we denote $O_{c, i}$ and $R_{c, i}$ as the transformed field at the capture plane $z_2$ for each time frame $i$. 

Then the captured intensity term $I$ can be expanded as $I = \vert O_c + R_c \vert^2 = O_c\cdot R_c^* + O_c^*\cdot R_c + O_c^2 + R_c^2$. (Time frame notation $i$ is omitted here.) In~\cite{Brady2009}, Brady et al. neglected the nonlinearity imposed by the squared magnitude and considered the two terms $O_c^2 + R_c^2$ (often referred to as noise and zero-order/DC term) as noise in the measurement model showing that they can be eliminated algorithmically using a CS reconstruction algorithm. In this work, we follow the same approach and the measured intensity can be expressed as
\begin{equation}
\label{eq:error}
I = \big\{ O_c\cdot R_c^* + O_c^*\cdot R_c \big\} +  O_c^2 + R_c^2 =2 Re\left\{ O_c\cdot R_c^*\right\} + E_E,
\end{equation}
where $E_E$ combines $O_c^2$ and $R_c^2$ into a single term considered as error. If we assume the object to be sparse, then the term $O_c^2$ is negligible. Thus, the error term can be approximately treated as $E_E \approx R_c^2$. In experiment, we approximate this error term by recording the background image and subtract the scene image by this background for reconstruction.

Now we assume that the sensor pixels have the same dimensions as the mask pixels, the unknown field $O$ will have spatial dimensions $N_{M_x} \times N_{M_y}$, depth dimension $N_d$ and temporal dimension $T$. Further, if we represent the convolutional operations in Eq.~\eqref{eq:discreteG} as Toeplitz matrices, we can obtain the following compact form
\begin{equation}
\label{eq:linear_model}
\begin{split}
{\bf g} &= 2S_TRe\Big\{ P_{T,h_2}\left\{ M_T \left[P_{T,h_1} \left(P_{T,d_n} {\bf o}\right)\right]\right\}\Big\} + {\bf e} + {\bf n} \\
& = A({\bf o}) + {\bf e} + {\bf n},
\end{split}
\end{equation}
where notation and dimensions of the introduced variables are summarized in Table~\ref{tab:variables} and $A(\cdot)$ describes the complete forward model.

In order to reconstruct the 4D volume, an optimization problem is formed as
\begin{equation} \label{eq:opt}
\hat{{\bf o}} = \displaystyle \argmin_{{\bf o}} \frac{1}{2}\left\lVert {\bf g} - A\left( {\bf o}\right) \right\rVert_2^2 + \lambda R({\bf o})
\end{equation} 
where $\lambda>0$ is a regularization parameter and $R(\cdot)$ is a regularizer on the unknown 4D field ${\bf o}$. 

In this work, we employ Total-Variation (TV) as the regularization function defined as
\begin{equation}\label{eq:TV}
R({\bf o}) = \left\lVert {\bf o} \right\rVert_{TV} = \sum_{t =0}^{T-1} \sum_{n = 1}^{N_d} \sum_{x = 1}^{N_{M_x}} \sum_{y=1}^{N_{M_y}}\left\vert \nabla\left( O\right)_{x,y,n,t} \right\vert,
\end{equation}
where we note here that ${\bf o}$ is the vectorized version of the unknown 4D object field $O: N_{M_x}\times N_{M_y} \times N_{d} \times T$. Eq.~\ref{eq:TV} is a generalized 4D TV regularizer. However, the choice of regularizer may vary by different purposes of reconstruction and/or properties of scenes. In experiment, a 3D TV (x, y, z) is used for resolving depths (Fig.~\ref{depth}). TV on temporal domain is included for recovering subtle movement (Fig.~\ref{fig:moving_fur}).  Also note that independent regularization parameters may be chosen for the spatial (x, y, z) and time (t) dimensions. We used Two-step IST (TwIST) algorithm~\cite{Bioucas2007} for reconstruction.

\begin{table}[!t]
\centering
\caption{Analysis of all the variables appearing in Eq.~\eqref{eq:linear_model}.}
\label{tab:variables}
\begin{tabular}{c}
\centering
$\begin{array}{l|l|l}
\textnormal{Variable} & \textnormal{Description} & \textnormal{Dimensions}\\ \hline
{\bf g}  & \textnormal{Vectorization of measured intensity $G$ from Eq.~\eqref{eq:error}.}      & (N_{M_x}\cdot N_{M_y}) \times 1\\ \hline

{\bf o}  &  \textnormal{Vectorization of unknown 4D object field $O$.}                     &(N_{M_x}\cdot N_{M_y}\cdot N_d \cdot T) \times 1 \\ \hline

{\bf e}  &  \textnormal{Vectorization of $E_E$ from Eq.~\eqref{eq:error}.} &(N_{M_x}\cdot N_{M_y}) \times 1\\ \hline

{\bf n}  &  \textnormal{Additive measurement noise vector.}                        &(N_{M_x}\cdot N_{M_y}) \times 1\\ \hline

P_{T,d_n} & 
\hspace{-1.5mm}\begin{array}{l}{\textnormal{Toeplitz matrix referring to $H_{d_n}$ from Eq.~\eqref{eq:discreteG}.}}\\{\textbf{\emph{Propagation and summation in depth (over $N_d$) for}}}\\{\textbf{\emph{all time frames.}}}\end{array}  
& \hspace{-1.5mm}\begin{array}{l}(N_{M_x}\cdot N_{M_y}\cdot T) \times\\ (N_{M_x}\cdot N_{M_y}\cdot N_d \cdot T)\end{array} \\ \hline

P_{T,h_1}& 
\hspace{-1.5mm}\begin{array}{l}{\textnormal{Toeplitz matrix referring to $H_{z_1 - z_0}$ from Eq.~\eqref{eq:discreteG}.}}\\{\textbf{\emph{Propagation of all time frames from $z_0$ to $z_1$.}}}\end{array}  
& \hspace{-1.5mm}\begin{array}{l}(N_{M_x}\cdot N_{M_y}\cdot T) \times\\ (N_{M_x}\cdot N_{M_y} \cdot T)\end{array} \\ \hline

M_T& 
\hspace{-1.5mm}\begin{array}{l}{\textnormal{Diagonal matrix containing all masks $M(x,y,t_i)$ for}}\\{\textnormal{all $i = 0 \dots T-1$ according to Eq.~\eqref{eq:discreteG}.}} \\ \textbf{\emph{Modulation of all time frames with different masks.}} \end{array}  
& \hspace{-1.5mm}\begin{array}{l}(N_{M_x}\cdot N_{M_y}\cdot T) \times\\ (N_{M_x}\cdot N_{M_y} \cdot T)\end{array} \\ \hline

P_{T,h_2} & 
\hspace{-1.5mm}\begin{array}{l}{\textnormal{Toeplitz matrix referring to $H_{z_2 - z_1}$ from Eq.~\eqref{eq:discreteG}.}}\\{\textbf{\emph{Propagation of all time frames from $z_1$ to $z_2$.}}}\end{array}  
& \hspace{-1.5mm}\begin{array}{l}(N_{M_x}\cdot N_{M_y}\cdot T) \times\\ (N_{M_x}\cdot N_{M_y} \cdot T)\end{array} \\ \hline

S_T& 
\hspace{-1.5mm}\begin{array}{l}{\textnormal{Matrix referring to the outer summation of Eq.~\eqref{eq:discreteG}.}}\\{\textbf{\emph{Summation in time (over $T$).}}}\end{array}  
& \hspace{-1.5mm}\begin{array}{l}(N_{M_x}\cdot N_{M_y}) \times\\ (N_{M_x}\cdot N_{M_y} \cdot T)\end{array} \\ \hline
\end{array}$
\end{tabular}
\end{table}

\section{Experimental}
\subsection{Setup}
\begin{figure}[!t]
	\centering
	\includegraphics[width=4.5in,height = 2.05in]{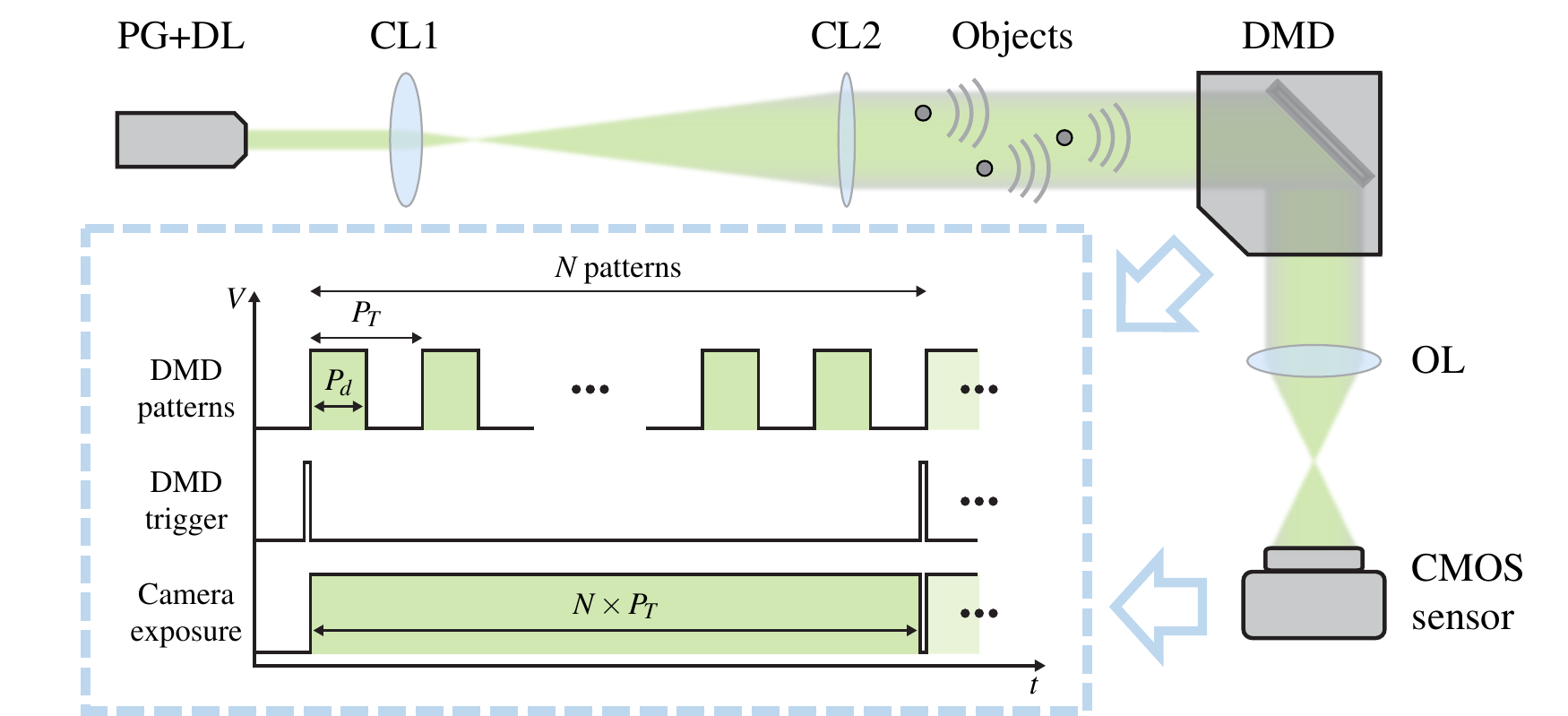}
	\caption{Schematic of the experimental setup. (PG: Pulse Generator; DL: Diode Laser; CL: Collimating Lens; DMD: Digital Micromirror Device; OL: Objective Lens. A trigger signal generated from the DMD is sent to the camera for exposure. The minimum time between successive DMD mask patterns is $P_T = 500 \mu s$ with a pattern exposure $P_d = 250 \mu s$. The camera is triggered every N patterns.}
	\label{fig:holo_experimental_setup}
\end{figure}

Fig.~\ref{fig:holo_experimental_setup} shows the schematic of the experimental setup. The illumination is produced by a diode laser with wavelength of 532 nm. The input beam is expanded and collimated by an ND filter and a collimating lens set (plano-convex lens, $300mm / 35mm = 8.57$ magnification, ND filter omitted). A digital micromirror device (DMD) is used to perform pixel-wise temporal modulation of the light field. For our experiments, we used the DLP\textregistered LightCrafter 4500\texttrademark from Texas Instruments Inc. The light engine includes a 0.45-inch DMD with $>1$million mirrors, each 7.6 $\mu m$, arranged in 912 columns by 1140 rows in a diamond pixel array geometry ~\cite{DMD}. The DMD is placed ~70mm distance away from the objects. An objective lens (single lens) is placed in front of the CMOS sensor and well-aligned with the DMD so that it images the DMD plane onto the sensor. The lens introduces a quadratic phase factor inside the integral of Eq. \ref{eq:propagation}. Thus, if the sensor is placed a distance of $2f$ from the OL, the phase is the same as $-2f$ from the lens. In this way, $T_{h_2}$ from Eq. \ref{eq:opt} reduces to the identity matrix. We used a CMOS monochromatic sensor with a resolution of $1920\times1200$ with a pixel pitch of $5.86 \mu m$. The key factor is the synchronization between the DMD and the sensor. Each DMD pattern can be projected as fast as $P_T=500 \mu s$ with an effective pattern exposure of $P_d=250 \mu s$. After N patterns are projected, a trigger signal is sent out to the camera which controls the shutter and results in a single exposure. 

\subsection{Subsampling holograms}
\begin{figure}[!t]
	\centering
	\includegraphics[width=5in,height=2.88in]{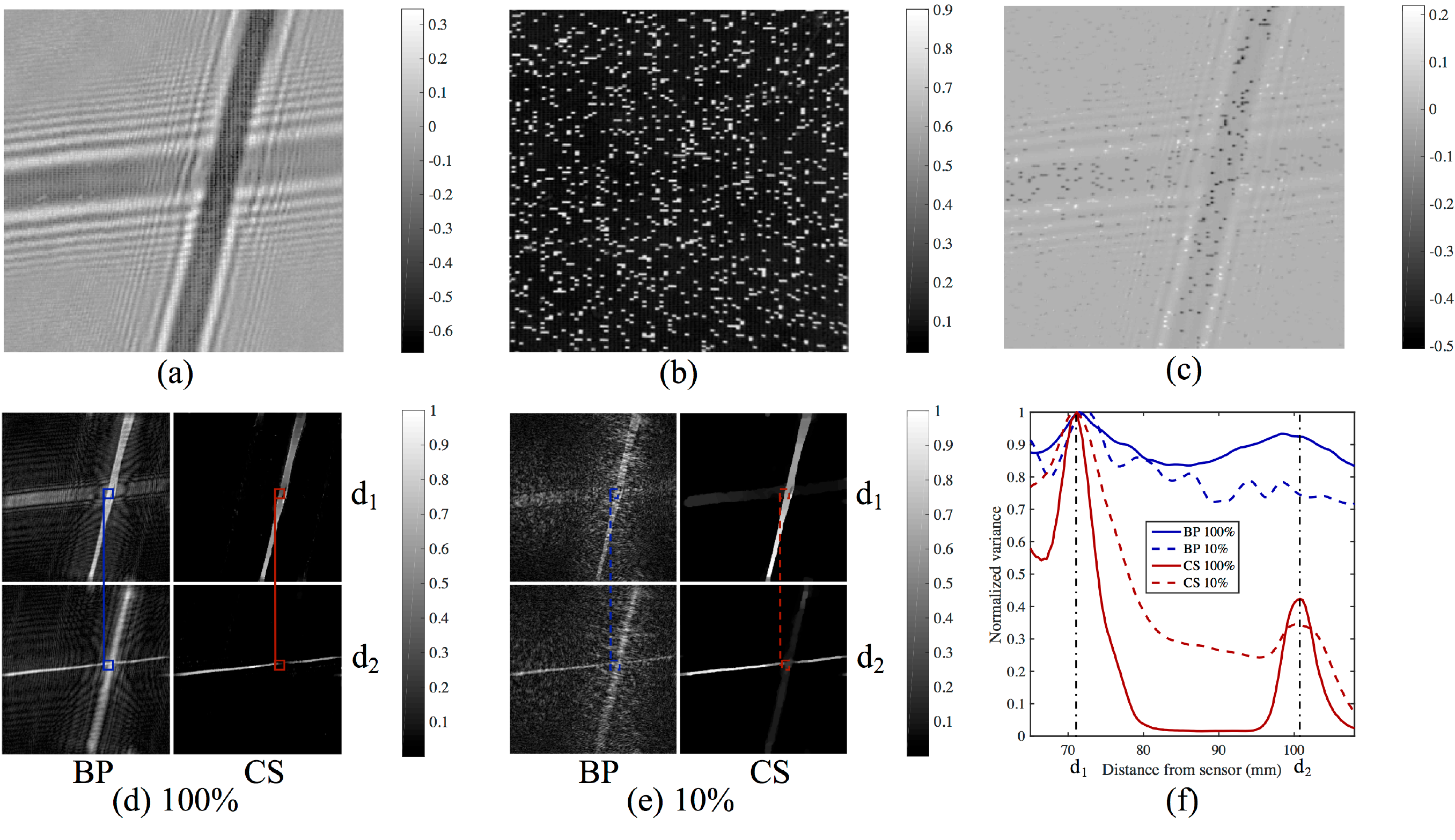}
	\caption{Subsampling holograms (background subtracted). (a)hologram of two static furs 7.1 cm and 10.1 cm away from sensor. (b) DMD mask, 10\%, uniformly random (background divided). (c) subsampled hologram. (d) Comparison of reconstructions from both back-propagation (BP) method and compressed sensing (CS) method using the full hologram. (e) Comparison for BP and CS using 10\% subsampled hologram. (f) Normalized variance vs. distance on z direction. Blue series: BP; red series: CS; full curve: 100\% hologram; dashed curve: 10\% hologram.}
	\label{depth}
\end{figure}

We start our experiment by examining the reconstruction performance of subsampled holograms. Recovery of a 3D object field from a 2D hologram has been proposed in previous work \cite{Brady2009}. The recovery can be treated as inference of high-dimensional data from undersampled measurements. Fig.~\ref{depth} shows the experimental results of 3D recovery with pixel-wise subsampling. For this experiment, we captured two static hairs from craft fur (see Fig.~\ref{fig:moving_fur}a) placed a distance of 7.1 cm and 10.1 cm away from the sensor. Fig.~\ref{depth} (a) shows the captured image. To preprocess the captured hologram, first we capture an image on the sensor with no object placed in the field of view - we refer to this as the background image. Note that this captured image corresponds to the term $R_c^2$ in Eq.~\ref{eq:error}. We then subtract the hologram by the background image, down-sampled to $960\times600$ and cropped the central $285\times285$ ROI around the object. Fig.~\ref{depth} (b) shows the captured image of one pattern from the DMD. Each pattern randomly selects 10\% of the entire image. To avoid aliasing artifacts caused by the diamond shaped sampling patterns on the DMD, we group together $4\times 4$ adjacent pixels on the DMD to make a single superpixel \cite{DMD}. In our reconstructions, to form the matrix A from Eq.~\ref{eq:linear_model}, we capture images of the mask with no object present. These captured images are divided by the background image to remove the effect of beam non-uniformity. Fig.~\ref{depth} (c) shows the subsampled hologram. Fig.~\ref{depth} (d) \& (e) compares reconstructions for the full hologram and subsampled hologram. The image ($285\times285$) was reconstructed into a 3D volume ($285\times285\times120$) with a depth range from 65 mm to 108 mm. Shown are the images reconstructed at the depth planes corresponding to the location of the two hairs. In order to quantify the performance in terms of depth resolution, we used block variance \cite{Mcelhinney2007} for the edge pixel of the cross section by the two hairs. Higher variance infers higher contrast, and thus, higher resolution. The block variance was computed within a window of $21\times21$ pixels highlighted as blue and red in Fig.~\ref{depth} (d) \& (e). Fig.~\ref{depth} (f) shows the normalized variance versus depth from sensor. Two principle peaks are observed and can be inferred as the focus distance for the two furs. The peak around $d_1$ has strong signal in all four curves. This was because the object located there has larger size than the other one. As can be seen, using only 10\% of the data deteriorates both BP and CS reconstruction resolutions. And in 10\% from BP, it is even harder to track the second object because of the impact of mask pattern. This can also be observed in the left panel of (e) where the back propagation of the mask severely affected the objects. The variance decreases fast in CS reconstructions. This implies the denoising effect as well as the optical sectioning power of CS. In 10\% reconstruction, the intermediate volume between the two objects were not denoised as good as in "100\%" case. This shows that greater subsampling factors reduce the effective depth resolution. 

\subsection{Temporal multiplexing}
\begin{figure}[!t]
	\centering
	\includegraphics[width=5in,height=3.11in]{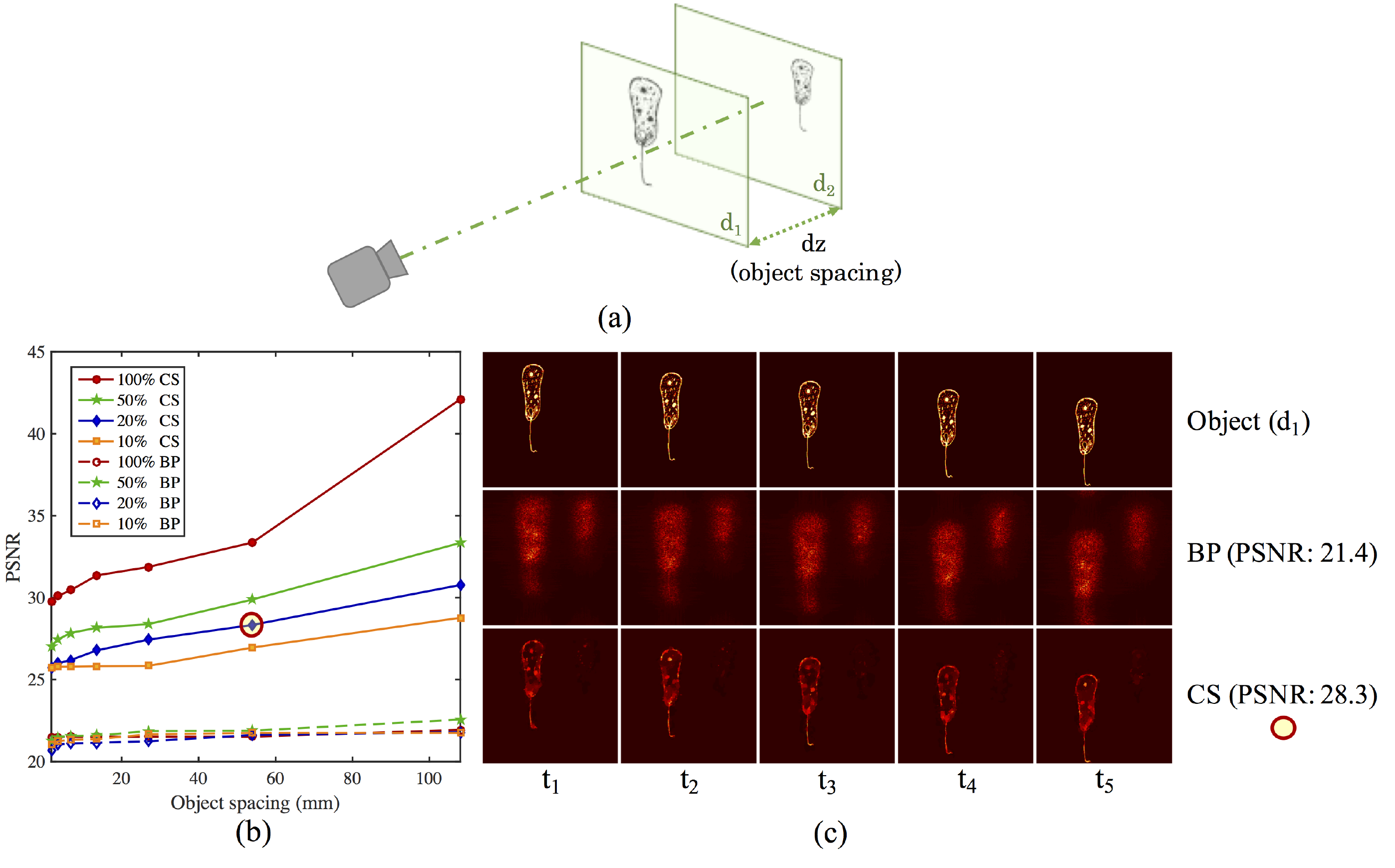}
	\caption{Simulated reconstruction performance for compressive holographic video. (a) simulation setup; (b) joint resolution (spatial and temporal). PSNR in dB. Lines represent CS results and dashed lines represent BP results. "100\%": full resolution; "50\%": temporal increase of 2; "20\%": temporal increase of 5; "10\%": temporal increase of 10. (c) reconstruction results at depth $d_1$. Marked as red circle in (b).}
	\label{fig:2dpsnr}
\end{figure}

In the previous section, we analyzed the effect of subsampling on reconstruction performance for compressive holography. Here we show how to utilize the excess pixel bandwidth in the sensor to increase temporal resolution. A simulation experiment was carried out in order to quantitatively analyze our imaging system (Fig.~\ref{fig:2dpsnr}). As shown in Fig.~\ref{fig:2dpsnr}(a), two layers of objects (peranema with different scales) were used as a test case. Each layer had $256\times256$ pixels. The pixel pitch was set so that the whole scene size  ($9.85\times9.85 mm$) was approximately identical to the DMD size. The first object was placed at $70 mm$ away from the sensor. The other object was placed $dz$ further away from the first object. $dz$ is a changing variable. A spatiotemporal subsampling mask was displayed on the DMD. For example, when $n$ time frames are required, each frame will have $1/n\times100\%$ of the pixels being randomly selected and displayed. In this way, the summation of $n$ frames is the full resolution scene image. In simulation, we omitted the propagation between the DMD and the sensor. For reconstruction, we compared back-propagation and compressed sensing. In order to have a better reconstruction result, we inserted 4 intermediate planes between the two objects. The results are shown in Fig.~\ref{fig:2dpsnr}(b) and (c). In (b), the peak signal-to-noise ratio (PSNR) was used to measure the reconstruction performance. $PSNR = 10log_{10}(peak value^2/MSE)$, where $MSE$ is the mean-squared error between the reconstruction and the input object field. The PSNR is computed on the 4D volume, which can also be treated as an average over multiple time frames. The higher PSNR value is, the better fidelity the reconstruction is. We picked out a point from (b), marked as red ring, to show in (c) the visual meaning of the PSNR values. It can be seen that lower rate of subsampling causes worse reconstruction performance. PSNR also decreases with the decrease of object spacing.

\subsection{Spatiotemporal recovery for fast-moving objects}
We present two illustrative examples which are aimed for the application of observing subtle vibrations and tracking small-but-fast-moving particles.

\begin{figure}[!t]
	\centering
	\includegraphics[width=5in,height=1.97in]{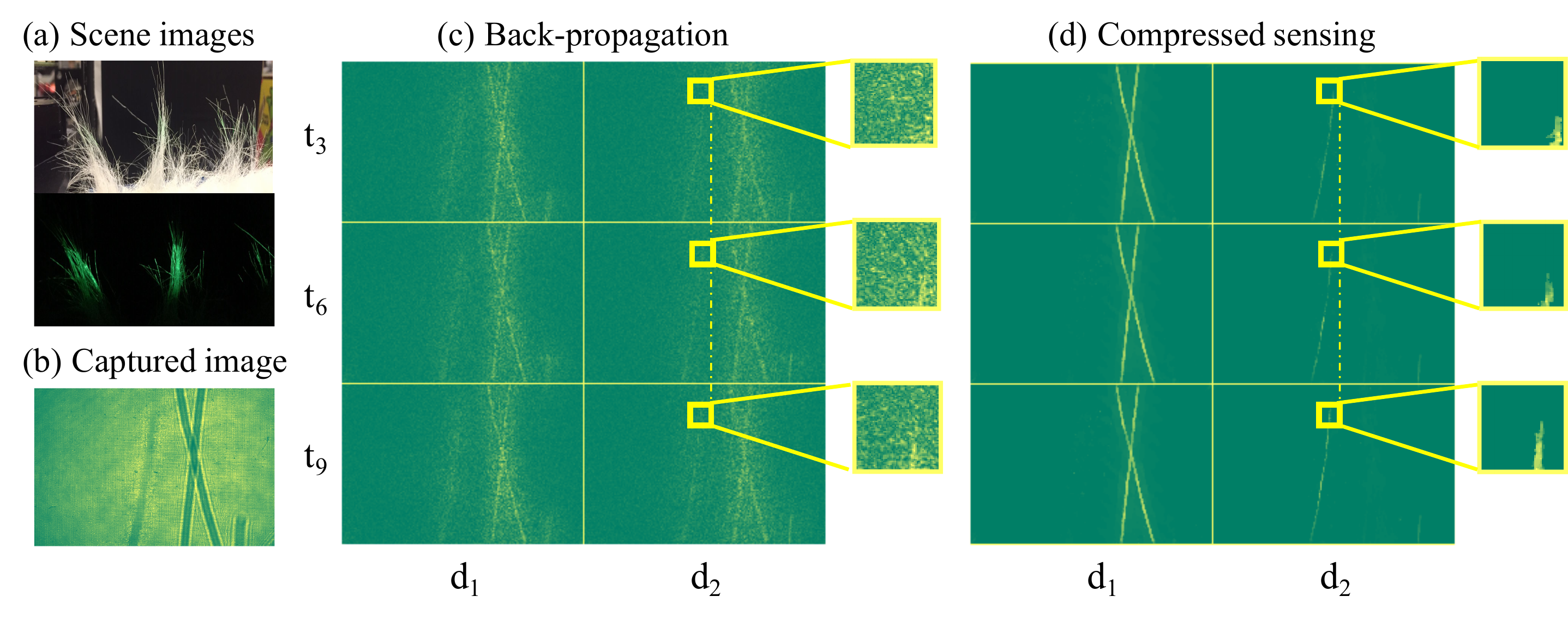}
	\caption{Reconstruction results from a single image (moving hairs). 10 frames of video and two depth frames are reconstructed from a single captured hologram. Due to space constraints, 3 video frames (3rd, 6th, 9th) and two depths ($d_1 = 73 mm, d_2 = 111 mm$) are presented. (Supplemental videos included.)}
	\label{fig:moving_fur}
\end{figure}

Fig.~\ref{fig:moving_fur} shows a reconstruction result demonstrating a $10\times$ increase in temporal resolution. The captured image contains several strands of hair blown by an air conditioner. \emph{From a single captured image, we reconstruct 2 depth slices and 10 frames of video}. In the case of small lateral movement, i.e. vibration, it is feasible to apply total variation on time domain. For the convenience of comparison, 3 time frames (3rd, 6th, 9th) are shown for both back-propagation and compressed sensing. In terms of depth, our CS result shows well-separated objects at different depth layers while the back-propagation method fails to achieve optical sectioning. The movement of the object is also recovered in our CS result.

\begin{figure}[!t]
	\centering
	\includegraphics[width=4.7in,height=4.91in]{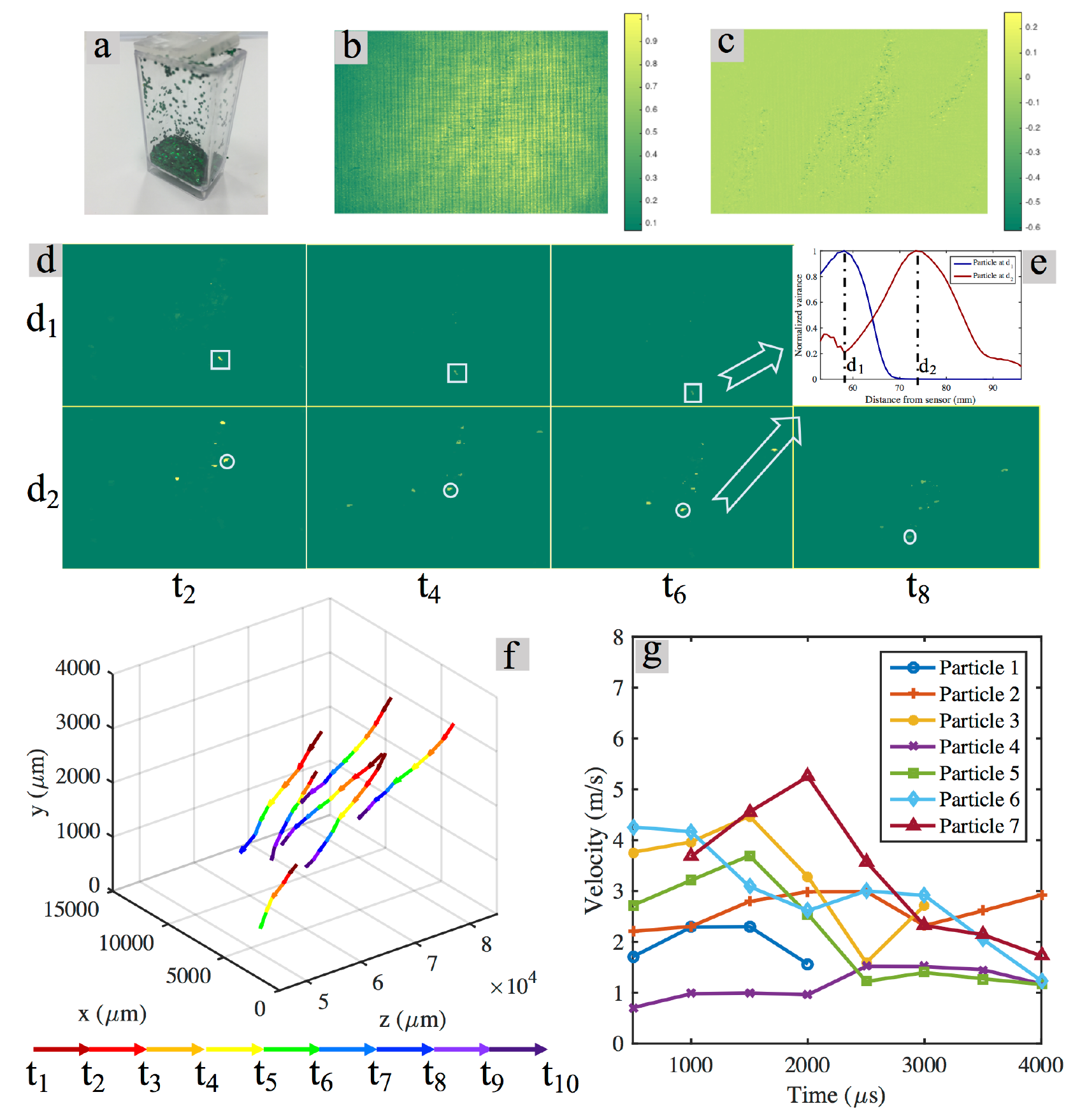}
	\caption{Reconstruction results from a single image (dropping glitters). (a) Glitters; (b) captured image; (c) normalized image; (d) reconstruction map. 2 depths and 4 out of 10 frames are shown; (e) normalized variance plot from 2 particles at $d_1$ and $d_2$; (f) 4D particle tracking; (g) velocity plotting with time range from $500 \mu s$ to $4000\mu s$.}
	\label{fig:glitters}
\end{figure}

Fig.~\ref{fig:glitters} shows another reconstruction result for dropping several flakes of glitter. The glitter flakes in Fig.~\ref{fig:glitters} (a) had size of 1 mm and were dropped in a range of 60 mm to 80 mm away from sensor. The glitter flakes were also blown by an air conditioner. (b) shows the captured single image. (c) shows preprocessed image which is subtracted by background image. In this case, the glitter flakes were moving at high speed. There was no overlap between two consecutive frames for the same flake. So the each frame was recovered independently. (d) shows a reconstruction map of 2 depths and 4 time frames. The downward and leftward motion of two glitter flakes can be observed. A similar refocusing method was used as in \cite{Mcelhinney2007}. Here, we scanned the reconstructed image by a $21\times21$ window and computed the variance (normalized) to get the focused depth information. If the normalized variance at defocused depth are higher than 0.5, that pixel was rejected as background/noise. For adjacent pixels which have similar variance profile, the pixels were treated as a single particle. (e) shows the normalized variance for two particles at $d_1$ and $d_2$. The particles are tracked at two locations pointed out by the arrows in (d). The overall tracking results are shown in Fig.~\ref{fig:glitters}(f) and (g). 7 particles are detected with 4D motion within 5 ms. In (f), the temporal transition was represented by 10 different color arrows. (g) shows a velocity chart of the 7 particles. The velocity of each particle was computed by $v(t_n) = [d(t_{n+1}) - d(t_{n-1})]/2\Delta t$, where $d(t_n)$ depicts the 3D location at $n$-th time frame, $\Delta t = 500 \mu s$. The velocity of the particles ranges from 0.7 m/s to 5.5 m/s.

\section{Conclusion}
We have demonstrated two illustrative cases where 4D spatio-temporal data is recovered from a single captured 2D hologram. In the case of vibrating hairs, 2 depth layers and 10 video frames in time were recovered. In the case of dropping glitter flakes, a 4D volume was reconstructed to track the motion of small particles. The prototype showed that it is possible to simultaneously exceed the capture rate of imagers and recover multiple depths with reasonable depth resolution. The coded-exposure technique enables high speed imaging with a simple frame rate camera. Digital in-line holography brings the capability of 3D tomographic imaging with simple experimental setup. Our Compressive Holographic Video technique is also closely related to phase retrieval problems commonly faced in holographic microscopy. Our space-time subsampling technique can be viewed as a sequence of coded apertures applied to a spatiotemporally varying optical field. In the future we plan to explore the connections between our CS reconstruction approach and the methods introduced in \cite{Horisaki2014}.

\section*{Acknowledgement}
The authors were grateful for the constructive discussions with Dr. Roarke Horstmeyer and Donghun Ryu. This work was funded in part by NSF CAREER grant IIS-1453192, ONR grant 1(GG010550)//N00014-14-1-0741, and ONR grant \#N00014-15-1-2735.
\end{document}